\documentclass[11pt,a4paper]{article}
\usepackage{authblk}
\usepackage[hyperref]{acl2020}
\usepackage{times}
\usepackage{latexsym}
\usepackage{bbding}
\usepackage{pifont}
\usepackage{verbatim}
\usepackage{booktabs,calc,scalerel}
\usepackage{amsmath,amssymb,stmaryrd}  
\usepackage{MnSymbol}
\usepackage{graphicx}

\usepackage{url}

\usepackage[splitrule]{footmisc}
\usepackage[compact]{titlesec}
\usepackage{microtype}

\aclfinalcopy 

\title{ERASER\,\scalerel*{\includegraphics{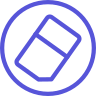}}{T}\,: A Benchmark to Evaluate Rationalized NLP Models}

\author[$\star\Psi$]{Jay DeYoung}
\author[$\star\Psi$]{Sarthak Jain}
\author[$\star\Phi$]{Nazneen Fatema Rajani}
\author[$\Psi$]{Eric Lehman}
\author[$\Phi$]{Caiming Xiong} 
\author[$\Phi$]{\authorcr Richard Socher}
\author[$\Psi$]{Byron C. Wallace}
\affil[$\star$]{Equal contribution.}
\affil[$\Psi$]{Khoury College of Computer Sciences, Northeastern University}
\affil[$\Phi$]{Salesforce Research, Palo Alto, CA, 94301}

\aclfinalcopy

\date{}

\begin{document}
\maketitle
\begin{abstract}
State-of-the-art models in NLP are now predominantly based on deep neural networks that are opaque in terms of how they come to make predictions. This limitation has increased interest in designing more interpretable deep models for NLP that reveal the `reasoning' behind model outputs. But work in this direction has been conducted on different datasets and tasks with correspondingly unique aims and metrics; this makes it difficult to track progress. 
We propose the {\bf E}valuating {\bf R}ationales {\bf A}nd {\bf S}imple {\bf E}nglish {\bf R}easoning ({\bf ERASER}\,\scalerel*{\includegraphics{figures/eraser.png}}{A}) benchmark to advance research on interpretable models in NLP. This benchmark comprises multiple datasets and tasks for which human annotations of ``rationales'' (supporting evidence) have been collected. We propose several metrics that aim to capture how well the rationales provided by models align with human rationales, and also how \emph{faithful} these rationales are (i.e., the degree to which provided rationales influenced the corresponding predictions). Our hope is that releasing this benchmark facilitates progress on designing more interpretable NLP systems. The benchmark, code, and documentation are available at \url{https://www.eraserbenchmark.com/}
\end{abstract}

\section{Introduction}

  \begin{figure}
 \centering
 \includegraphics[scale=.4]{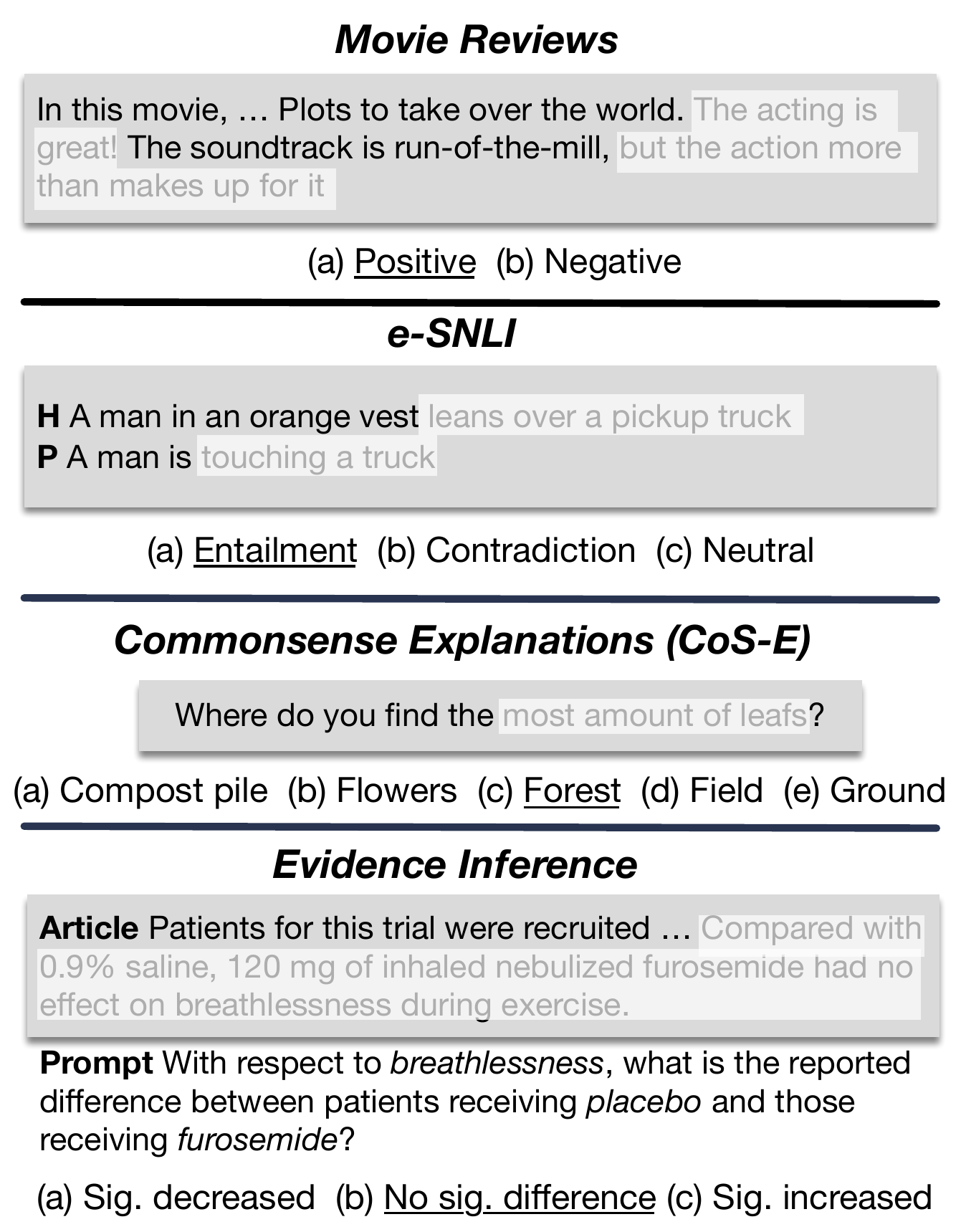}
\caption{Examples of instances, labels, and rationales illustrative of four (out of seven) datasets included in ERASER. The `erased' snippets  are rationales.} \label{eraser_annotations}
 \end{figure}
 

Interest has recently grown in designing NLP systems that can reveal \textbf{why} models make specific predictions.
But work in this direction has been conducted on different datasets and using different metrics to quantify performance; this has made it difficult to compare methods and track progress.
We aim to address this issue by releasing a standardized benchmark of datasets --- repurposed and augmented from pre-existing corpora, spanning a range of NLP tasks --- and associated metrics for measuring different properties of rationales.
We refer to this as the {\bf E}valuating {\bf R}ationales {\bf A}nd {\bf S}imple {\bf E}nglish {\bf R}easoning ({\bf ERASER}\,\scalerel*{\includegraphics{figures/eraser.png}}{A}) benchmark.

In curating and releasing ERASER we take inspiration from the stickiness of the GLUE~\citep{wang2018glue} and SuperGLUE \cite{DBLP:journals/corr/abs-1905-00537} benchmarks for evaluating progress in natural language understanding tasks, which have driven rapid progress on models for general language representation learning.
We believe the still somewhat nascent subfield of interpretable NLP stands to benefit similarly from an analogous collection of standardized datasets and tasks; we hope these will aid the design of standardized metrics to measure different properties of `interpretability', and we propose a set of such metrics as a starting point. 


Interpretability is a broad topic with many possible realizations \cite{doshi2017towards,lipton2016mythos}. In ERASER we focus specifically on \emph{rationales}, i.e., snippets that support outputs.
All datasets in ERASER include such rationales, explicitly marked by human annotators. 
By definition, rationales should be \emph{sufficient} to make predictions, but they may not be \emph{comprehensive}. Therefore, for some datasets, we have also collected comprehensive rationales (in which \emph{all} evidence supporting an output has been marked) on test instances.

The `quality' of extracted rationales will depend on their intended use. Therefore, we propose an initial set of metrics to evaluate rationales that are meant to measure different varieties of `interpretability'. Broadly, this includes measures of agreement with human-provided rationales, and assessments of \emph{faithfulness}. The latter aim to capture the extent to which rationales provided by a model in fact informed its predictions. 
We believe these provide a reasonable start, but view the problem of designing metrics for evaluating rationales --- especially for measuring faithfulness --- as a topic for further research that ERASER can facilitate. 
And while we will provide a `leaderboard', this is better viewed as a `results board'; we do not privilege any one metric. Instead, ERASER permits comparison between models that provide rationales with respect to different criteria of interest. 

We implement baseline models and report their performance across the corpora in ERASER. We find that no single `off-the-shelf' architecture is readily adaptable to datasets with very different instance lengths and associated rationale snippets (Section~\ref{section:datasets}). This highlights a need for new models that can consume potentially lengthy inputs and adaptively provide rationales at a task-appropriate level of granularity. ERASER provides a resource to develop such models.

In sum, we introduce the ERASER benchmark (\url{www.eraserbenchmark.com}), a unified set of diverse NLP datasets (these are repurposed and augmented from existing corpora,\footnote{We ask users of the benchmark to cite all original papers, and provide a BibTeX entry for doing so on the website.} including sentiment analysis, Natural Language Inference, and QA tasks, among others) in a standardized format featuring human rationales for decisions, along with starter code and tools, baseline models, and standardized (initial) metrics for rationales.
\section{Related Work} 

Interpretability in NLP is a large, fast-growing area; we do not attempt to provide a comprehensive overview here. Instead we focus on directions particularly relevant to ERASER, i.e., prior work on models that provide rationales for their predictions.

\vspace{.25em}
\noindent {\bf Learning to explain}. In ERASER we assume that rationales (marked by humans) are provided during training. However, such direct supervision will not always be available, motivating work on methods that can explain (or ``rationalize'') model predictions using only instance-level supervision.

In the context of modern neural models for text classification, one might use variants of \emph{attention} \citep{bahdanau2014neural} to extract rationales. Attention mechanisms learn to assign soft weights to (usually contextualized) token representations, and so one can extract highly weighted tokens as rationales. However, attention weights do not in general provide faithful explanations for predictions \citep{jain2019attention,serrano2019attention,wiegreffe2019attention,zhong2019fine,pruthi2019learning,brunner2019validity,moradi2019interrogating,vashishth2019attention}. This likely owes to encoders entangling inputs, complicating the interpretation of attention weights on \emph{inputs} over \emph{contextualized representations} of the same.\footnote{Interestingly, \citet{zhong2019fine} find that attention sometimes provides \emph{plausible} but not \emph{faithful} rationales. Elsewhere, Pruthi \emph{et al.} \shortcite{pruthi2019learning} show that one can easily learn to deceive via attention weights. These findings highlight that one should be mindful of the criteria one wants rationales to fulfill.} 

By contrast, \emph{hard} attention mechanisms discretely extract snippets from the input to pass to the classifier, by construction providing faithful explanations. Recent work has proposed hard attention mechanisms as a means of providing explanations. \citet{lei2016rationalizing} proposed instantiating two models with their own parameters; one to extract rationales, and one that consumes these to make a prediction. They trained these models jointly via REINFORCE \citep{williams1992simple} style optimization. 

Recently, \citet{jain2020} proposed a variant of this two-model setup that uses heuristic feature scores to derive pseudo-labels on tokens comprising rationales; one model can then be used to perform hard extraction in this way, while a second (independent) model can make predictions on the basis of these. Elsewhere, ~\citet{chang2019game} introduced the notion of classwise rationales that explains support for different output classes using a game theoretic framework. Finally, other recent work has proposed using a differentiable binary mask over inputs, which also avoids recourse to REINFORCE \citep{bastings-etal-2019-interpretable}.

\vspace{.25em}
\noindent {\bf Post-hoc explanation}. Another strand of interpretability work considers \emph{post-hoc} explanation methods, which seek to explain why a model made a specific prediction for a given input. Commonly these take the form of token-level importance scores. Gradient-based explanations are a standard example \cite{sundararajan2017axiomatic,smilkov2017smoothgrad}. These enjoy a clear semantics (describing how perturbing inputs locally affects outputs), but may nonetheless exhibit counterintuitive behaviors \cite{feng2018pathologies}. 


Gradients of course assume model differentiability. Other methods do not require any model properties. Examples include LIME~\citep{ribeiro2016should} and ~\citet{alvarez2017causal}; these methods approximate model behavior locally by having it repeatedly make predictions over perturbed inputs and fitting a simple, explainable model over the outputs.



\vspace{.25em}
\noindent {\bf Acquiring rationales}. Aside from interpretability considerations, collecting rationales from annotators may afford greater efficiency in terms of model performance realized given a fixed amount of annotator effort \citep{zaidan2008modeling}. In particular, recent work by ~\citet{mcdonnell2017many,mcdonnell2016relevant} has observed that at least for some tasks, asking annotators to provide rationales justifying their categorizations does not impose much additional effort. Combining rationale annotation with \emph{active learning} \cite{settles2012active} is another promising direction \cite{wallace2010active,sharma2015active}.


\vspace{.25em}
\noindent{\bf Learning from rationales}. Work on learning from rationales marked by annotators for text classification dates back over a decade~\citep{zaidan2007using}. 
Earlier efforts proposed extending standard discriminative models like Support Vector Machines (SVMs) with regularization terms that penalized parameter estimates which disagreed with provided rationales \citep{zaidan2007using,small2011constrained}.
Other efforts have attempted to specify \emph{generative} models of rationales~\citep{zaidan2008modeling}. 

More recent work has aimed to exploit rationales in training neural text classifiers. \citet{zhang2016rationale} proposed a \emph{rationale-augmented} Convolutional Neural Network (CNN) for text classification, explicitly trained to identify sentences supporting categorizations.~\citet{strout2019human} showed that providing this model with rationales during training yields predicted rationales that are preferred by humans (compared to rationales produced without explicit supervision). 
Other work has proposed `pipeline' approaches in which independent models are trained to perform rationale extraction and classification on the basis of these, respectively \citep{lehman2019inferring,chen-etal-2019-seeing}, assuming explicit training data is available for the former.





\citet{rajani2019explain} fine-tuned a Transformer-based language model \citep{radford2018improving} on free-text rationales provided by humans, with an objective of generating open-ended explanations to improve performance on downstream tasks.

\vspace{.25em}
\noindent{\bf Evaluating rationales}. Work on evaluating rationales has often compared these to human judgments \cite{strout2019human,doshi2017towards}, or elicited other human evaluations of explanations \citep{ribeiro2016should, lundberg2017unified, nguyen2018comparing}. There has also been work on visual evaluations of saliency maps~\citep{li-etal-2016-visualizing, ding-etal-2017-visualizing, sundararajan2017axiomatic}. 

Measuring agreement between extracted and human rationales (or collecting subjective assessments of them) assesses the plausibility of rationales, but such approaches do not establish whether the model actually relied on these particular rationales to make a prediction. 
We refer to rationales that correspond to the inputs most relied upon to come to a disposition as \emph{faithful}.

Most automatic evaluations of faithfulness measure the impact of perturbing or erasing words or tokens identified as important on model output \citep{Arras2017WhatIR, montavon2017explaining, serrano2019attention, samek2016evaluating,jain2019attention}. We build upon these methods in Section \ref{section:metrics}. Finally, we note that a recent article urges the community to evaluate faithfulness on a continuous scale of acceptability, rather than viewing this as a binary proposition \citep{jacovi}.






\section{Datasets in ERASER}
\label{section:datasets}



\begin{table}
\centering
\scriptsize
\begin{tabular}{lccc}\toprule
\textbf{Name} & \textbf{Size (train/dev/test)} & \textbf{Tokens} & \textbf{Comp?} \\
\hline 
Evidence Inference & 7958 / 972 / 959  & 4761 & $\diamond$  \\
BoolQ & 6363 / 1491 / 2817 & 3583 & $\diamond$ \\ 
Movie Reviews & 1600 / 200 / 200  & 774 &$\filleddiamond$\\ 
FEVER  & 97957 / 6122 / 6111   & 327 & \Checkmark \\
MultiRC & 24029 / 3214 / 4848  & 303 & \Checkmark \\ 
CoS-E & 8733 / 1092 / 1092  & 28 & \Checkmark \\ 
e-SNLI & 911938 / 16449 / 16429 & 16 & \Checkmark \\
\bottomrule
\end{tabular}
\caption{Overview of datasets in the ERASER benchmark. \textit{Tokens} is the average number of tokens in each document. Comprehensive rationales mean that all supporting evidence is marked; \Checkmark denotes cases where this is (more or less) true by default; $\diamond$, $\filleddiamond$ are datasets for which we have collected comprehensive rationales for either a subset or all of the test datasets, respectively.
Additional information can be found in Appendix \ref{section:preprocessing}.}.
\vspace{-2em}
\end{table}\label{table:datasets}

For all datasets in ERASER we distribute both reference labels and rationales marked by humans as supporting these in a standardized format. We delineate train, validation, and test splits for all corpora (see Appendix \ref{section:preprocessing} for processing details). We ensure that these splits comprise disjoint sets of source documents to avoid contamination.\footnote{Except for BoolQ, wherein source documents in the original train and validation set were not disjoint and we preserve this structure in our dataset. \emph{Questions}, of course, are disjoint.}
We have made the decision to distribute the test sets publicly,\footnote{Consequently, for datasets that have been part of previous benchmarks with other aims (namely, GLUE/superGLUE) but which we have re-purposed for work on rationales in ERASER, e.g., BoolQ \citep{clark2019boolq}, we have carved out for release test sets from the original validation sets.} in part because we do not view the `correct' metrics to use as settled. We plan to acquire additional human annotations on held-out portions of some of the included corpora so as to offer hidden test set evaluation opportunities in the future.  



\vspace{.25em} \noindent {\bf Evidence inference} \citep{lehman2019inferring}. A dataset of full-text articles describing randomized controlled trials (RCTs). The task is to infer whether a given \emph{intervention} is reported to either \emph{significantly increase}, \emph{significantly decrease}, or have \emph{no significant effect} on a specified \emph{outcome}, as compared to a \emph{comparator} of interest. Rationales have been marked as supporting these inferences. As the original annotations are not necessarily exhaustive, we collected exhaustive rationale annotations on a subset of the validation and test data.\footnote{\label{appx-ann}Annotation details are in  Appendix~\ref{section:ann-details}.}

\vspace{.25em} \noindent {\bf BoolQ} \citep{clark2019boolq}. This corpus consists of passages selected from Wikipedia, and yes/no questions generated from these passages. As the original Wikipedia article versions used were not maintained, we have made a best-effort attempt to recover these, and then find within them the passages answering the corresponding questions. For public release, we acquired comprehensive annotations on a subset of documents in our test set.\footref{appx-ann}

\vspace{.25em} \noindent {\bf Movie Reviews} \citep{zaidan2008modeling}. Includes positive/negative sentiment labels on movie reviews. Original rationale annotations were not necessarily comprehensive; we thus collected comprehensive rationales on the final two folds of the original dataset \citep{pang-lee-2004-sentimental}.\footref{appx-ann} In contrast to most other datasets, the rationale annotations here are \textit{span level} as opposed to sentence level.

\vspace{.25em} \noindent {\bf FEVER} \citep{thorne2018fever}. Short for Fact Extraction and VERification; entails verifying claims from textual sources. Specifically, each claim is to be classified as \emph{supported}, \emph{refuted} or \emph{not enough information} with reference to a collection of source texts. We take a subset of this dataset, including only \emph{supported} and \emph{refuted} claims.

\vspace{.25em} \noindent {\bf MultiRC} \citep{KCRUR18-multirc}. A reading comprehension dataset composed of questions with multiple correct answers that by construction depend on information from multiple sentences. Here each rationale is associated with a question, while answers are independent of one another. We convert each rationale/question/answer triplet into an instance within our dataset. Each answer candidate then has a label of \emph{True} or \emph{False}.





\vspace{.25em} \noindent {\bf Commonsense Explanations (CoS-E)} \citep{rajani2019explain}. This corpus comprises multiple-choice questions and answers from \citep{talmor-etal-2019-commonsenseqa} along with supporting rationales. The rationales in this case come in the form both of highlighted (extracted) supporting snippets and free-text, open-ended descriptions of reasoning. Given our focus on extractive rationales, ERASER includes only the former for now. Following \citet{talmor-etal-2019-commonsenseqa}, we repartition the training and validation sets to provide a canonical test split.

\vspace{.25em} \noindent {\bf e-SNLI}~\citep{camburu2018snli}. This dataset augments the SNLI corpus~\citep{snli:emnlp2015} with rationales marked in the premise and/or hypothesis (and natural language explanations, which we do not use). For entailment pairs, annotators were required to highlight at least one word in the premise. For contradiction pairs, annotators had to highlight at least one word in both the premise and the hypothesis; for neutral pairs, they were only allowed to highlight words in the hypothesis.

\vspace{.25em} \noindent {\bf Human Agreement} We report human agreement over extracted rationales for multiple annotators and documents in Table \ref{table:human_agreement}. All datasets have a high Cohen $\kappa$ ~\citep{Cohen1960}; with substantial or better agreement.

\begin{table*}[ht]
\scriptsize
\centering
\begin{tabular}{lcccccc}
\toprule
     Dataset & Cohen $\kappa$ & F1 & P & R & \#Annotators/doc & \#Documents \\ \hline
     Evidence Inference & - & - & - & - & - & - \\
    BoolQ & 0.618 $\pm$ 0.194 & 0.617 $\pm$ 0.227 & 0.647 $\pm$ 0.260 & 0.726 $\pm$ 0.217 & 3 & 199 \\
     Movie Reviews & 0.712 $\pm$ 0.135 & 0.799 $\pm$ 0.138 & 0.693 $\pm$ 0.153 & 0.989 $\pm$ 0.102 & 2 & 96 \\
     FEVER & 0.854 $\pm$ 0.196 & 0.871 $\pm$ 0.197 & 0.931 $\pm$ 0.205 & 0.855 $\pm$ 0.198 & 2 & 24 \\
     MultiRC & 0.728 $\pm$ 0.268 & 0.749 $\pm$ 0.265 & 0.695 $\pm$ 0.284 & 0.910 $\pm$ 0.259 & 2 & 99 \\
     CoS-E & 0.619 $\pm$ 0.308 & 0.654 $ \pm$ 0.317 & 0.626 $\pm$ 0.319 & 0.792 $\pm$ 0.371 & 2 & 100 \\
     e-SNLI & 0.743 $\pm$ 0.162 & 0.799 $\pm$ 0.130 & 0.812 $\pm$ 0.154 & 0.853 $\pm$ 0.124 & 3 & 9807 \\
     \bottomrule
\end{tabular}
\caption{Human agreement with respect to rationales. For Movie Reviews and BoolQ we calculate the mean agreement of individual annotators with the majority vote per token, over the two-three annotators we hired via Upwork and Amazon Turk, respectively. The e-SNLI dataset already comprised three annotators; for this we calculate mean agreement between individuals and the majority. For CoS-E, MultiRC, and FEVER, members of our team annotated a subset to use a comparison to the (majority of, where appropriate) existing rationales. We collected comprehensive rationales for Evidence Inference from Medical Doctors; as they have a high amount of expertise, we would expect agreement to be high, but have not collected redundant comprehensive annotations.} 
\label{table:human_agreement}
\vspace{-2em}
\end{table*}
\section{Metrics}
\label{section:metrics} 

In ERASER models are evaluated both for their predictive performance and with respect to the rationales that they extract. For the former, we rely on the established metrics for the respective tasks. Here we describe the metrics we propose to evaluate the quality of extracted rationales. We do not claim that these are necessarily the best metrics for evaluating rationales, however. Indeed, we hope the release of ERASER will spur additional research into how best to measure the quality of model explanations in the context of NLP.

\subsection{Agreement with human rationales}

The simplest means of evaluating extracted rationales is to measure how well they agree with those marked by humans. We consider two classes of metrics, appropriate for models that perform discrete and `soft' selection, respectively. 



For the discrete case, measuring exact matches between predicted and reference rationales is likely too harsh.\footnote{Consider that an extra token destroys the match but not usually the meaning} We thus consider more relaxed measures. These include Intersection-Over-Union (IOU), borrowed from computer vision~\citep{Everingham2010}, which permits credit assignment for partial matches. We define IOU on a token level: for two spans, it is the size of the overlap of the tokens they cover divided by the size of their union. We count a prediction as a match if it overlaps with any of the ground truth rationales by more than some threshold (here, 0.5). We use these partial matches to calculate an F1 score. We also measure \emph{token}-level precision and recall, and use these to derive token-level F1 scores.




Metrics for continuous or soft token scoring models consider token rankings, rewarding models for assigning higher scores to marked tokens. In particular, we take the Area Under the Precision-Recall curve (AUPRC) constructed by sweeping a threshold over token scores. We define additional metrics for soft scoring models below.

In general, the rationales we have for tasks are \emph{sufficient} to make judgments, but not necessarily \emph{comprehensive}. However, for some datasets we have explicitly collected comprehensive rationales for at least a subset of the test set. Therefore, on these datasets \emph{recall} evaluates comprehensiveness directly (it does so only noisily on other datasets). We highlight which corpora contain comprehensive rationales in the test set in Table \ref{table:datasets}.

\subsection{Measuring faithfulness}
\label{subsection:faithfulness}

  \begin{figure*}
 \centering
 \includegraphics[width=0.8\textwidth]{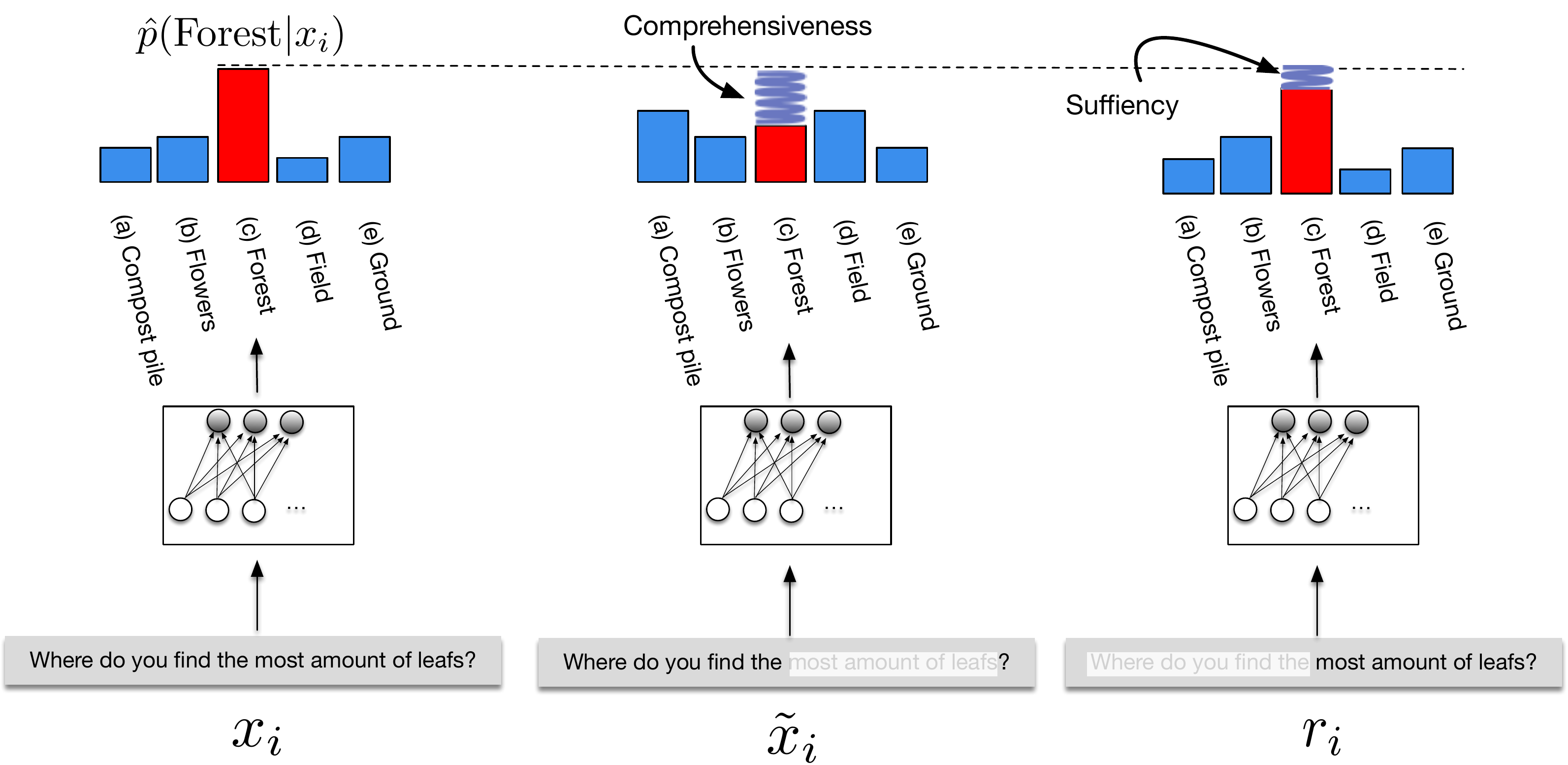}
 \vspace{-.5em}
\caption{Illustration of faithfulness scoring metrics, \emph{comprehensiveness} and \emph{sufficiency}, on the Commonsense Explanations (CoS-E) dataset. For the former, erasing the tokens comprising the provided rationale ($\tilde{x}_i$) ought to decrease model confidence in the output `Forest'. For the latter, the model should be able to come to a similar disposition regarding `Forest' using \emph{only} the rationales $r_i$.} \label{fig:faithful} 
 \vspace{-1.0em}
 \end{figure*}
 
As discussed above, a model may provide rationales that are plausible (agreeable to humans) but that it did not rely on for its output. In many settings one may want rationales that actually \emph{explain} model predictions, i.e., rationales extracted for an instance in this case ought to have meaningfully influenced its prediction for the same. We call these faithful rationales.
How best to measure rationale faithfulness is an open question. In this first version of ERASER we propose simple metrics motivated by prior work~\citep{zaidan2007using,yu2019rethinking}. In particular, following~\citet{yu2019rethinking} we define metrics intended to measure the \emph{comprehensiveness} (were all features needed to make a prediction selected?) and \emph{sufficiency} (do the extracted rationales contain enough signal to come to a disposition?) of rationales, respectively. 


{\bf Comprehensiveness}. To calculate rationale comprehensiveness we create \emph{contrast} examples~\citep{zaidan2007using}: We construct a contrast example for $x_i$, $\tilde{x}_i$, which is $x_i$ with the predicted rationales $r_i$ removed. Assuming a classification setting, let $m(x_i)_j$ be the original prediction provided by a model $m$ for the predicted class $j$. Then we consider the predicted probability from the model for the same class once the supporting rationales are stripped. Intuitively, the model ought to be less confident in its prediction once rationales are removed from $x_i$. We can measure this as: 
\begin{equation}
{\text{comprehensiveness}} = m(x_i)_j - m(x_i \backslash r_i)_j
\label{eq:comprehensiveness}
\end{equation}  

\noindent A high score here implies that the rationales were indeed influential in the prediction, while a low score suggests that they were not. A negative value here means that the model became \emph{more} confident in its prediction after the rationales were removed; this would seem counter-intuitive if the rationales were indeed the reason for its prediction.

{\bf Sufficiency}. This captures the degree to which the snippets within the extracted rationales are adequate for a model to make a prediction. 
\begin{equation}
{\text{sufficiency}} = m(x_i)_j - m(r_i)_j 
\label{eq:sufficiency}
\end{equation}  

\noindent These metrics are illustrated in Figure \ref{fig:faithful}.

As defined, the above measures have assumed discrete rationales $r_i$. We would also like to evaluate the faithfulness of continuous importance scores assigned to tokens by models. Here we adopt a simple approach for this. We convert soft scores over features $s_i$ provided by a model into discrete rationales $r_i$ by taking the top$-k_d$ values, where $k_d$ is a threshold for dataset $d$. We set $k_d$ to the average rationale length provided by humans for dataset $d$ (see Table~\ref{tab:soft-scores}). Intuitively, this says: How much does the model prediction change if we remove a number of tokens equal to what humans use (on average for this dataset) in order of the importance scores assigned to these by the model. Once we have discretized the soft scores into rationales in this way, we compute the faithfulness scores as per Equations \ref{eq:comprehensiveness} and \ref{eq:sufficiency}.

This approach is conceptually simple. It is also computationally cheap to evaluate, in contrast to measures that require per-token measurements, e.g., importance score correlations with `leave-one-out' scores~\citep{jain2019attention}, or counting how many `important' tokens need to be erased before a prediction flips~\citep{serrano2019attention}. However, the necessity of discretizing continuous scores forces us to pick a particular threshold $k$. 

We can also consider the behavior of these measures as a function of $k$, inspired by the measurements proposed in \citet{samek2016evaluating} in the context of evaluating saliency maps for image classification. They suggested ranking pixel regions by importance and then measuring the change in output as they are removed in rank order. Our datasets comprise documents and rationales with quite different lengths; to make this measure comparable across datasets, we construct bins designating the number of tokens to be deleted. Denoting the tokens up to and including bin $k$ for instance $i$ by $r_{ik}$, we define an aggregate comprehensiveness measure:

\begin{equation}
 \frac{1}{|\mathcal{B}|+1} (\sum_{k=0}^{|\mathcal{B}|} m(x_i)_j -  m(x_i \backslash r_{ik})_j) 
    \label{equation:AOPC}
\end{equation}

This is defined for sufficiency analogously. Here we group tokens into $k=5$ bins by grouping them into the top 1\%, 5\%, 10\%, 20\% and 50\% of tokens, with respect to the corresponding importance score. We refer to these metrics as ``Area Over the Perturbation Curve" (AOPC).\footnote{Our AOPC metrics are similar in concept to ROAR \cite{hooker2019} except that we re-use an existing model as opposed to retraining for each fraction.}

These AOPC sufficiency and comprehensiveness measures score a particular token ordering under a model. As a point of reference, we also report these when \emph{random} scores are assigned to tokens. 

\section{Baseline Models} 
\label{section:baselines}

Our focus in this work is primarily on the ERASER benchmark itself, rather than on any particular model(s). 
But to establish a starting point for future work, we evaluate several baseline models across the corpora in ERASER.\footnote{This is not intended to be comprehensive.}
We broadly classify these into models that assign `soft' (continuous) scores to tokens, and those that perform a `hard' (discrete) selection over inputs. We additionally consider models specifically designed to select individual tokens (and very short sequences) as rationales, as compared to longer snippets. 
All of our implementations are in PyTorch ~\citep{paszke2019pytorch} and are available in the ERASER repository.\footnote{\url{https://github.com/jayded/eraserbenchmark}}

All datasets in ERASER comprise inputs, rationales, and labels. But they differ considerably in document and rationale lengths (Table \ref{table:detailed_dataset}).
This motivated use of different models for datasets, appropriate to their sizes and rationale granularities. We hope that this benchmark motivates design of models that provide rationales that can flexibly adapt to varying input lengths and expected rationale granularities. Indeed, only with such models can we perform comparisons across all datasets.


\subsection{Hard selection}

Models that perform \emph{hard} selection may be viewed as comprising two independent modules: an \emph{encoder} which is responsible for extracting snippets of inputs, and a \emph{decoder} that makes a prediction based only on the text provided by the encoder. We consider two variants of such models.

\vspace{.25em}
\noindent{\bf\citet{lei2016rationalizing}}. In this model, an encoder induces a binary mask over inputs $x$, $z$. The decoder accepts the tokens in $x$ unmasked by $z$ to make a prediction $\hat{y}$. These modules are trained jointly via REINFORCE~\citep{williams1992simple} style estimation, minimizing the loss over expected binary vectors $z$ yielded from the encoder. One of the advantages of this approach is that it need not have access to marked rationales; it can learn to rationalize on the basis of instance labels alone. However, given that we do have rationales in the training data, we experiment with a variant in which we train the encoder explicitly using rationale-level annotations.

In our implementation of ~\citet{lei2016rationalizing}, we drop in two independent BERT~\citep{devlin2018bert} or GloVe~\citep{pennington-etal-2014-glove} base modules with bidirectional LSTMs~\citep{hochreiter1997long} on top to induce contextualized representations of tokens for the encoder and decoder, respectively. The encoder generates a scalar (denoting the probability of selecting that token) for each LSTM hidden state using a feedfoward layer and sigmoid. In the variant using human rationales during training, we minimize cross entropy loss over rationale predictions. The final loss is then a composite of classification loss, regularizers on rationales \cite{lei2016rationalizing}, and loss over rationale predictions, when available. 


\vspace{.25em}
\noindent {\bf Pipeline models}. These are simple models in which we first train the encoder to extract rationales, and then train the decoder to perform prediction using only rationales. No parameters are shared between the two models. 

Here we first consider a simple pipeline that first segments inputs into sentences. It passes these, one at a time, through a Gated Recurrent Unit (GRU)~\citep{cho2014learning}, to yield hidden representations that we compose via an attentive decoding layer ~\citep{bahdanau2014neural}. This aggregate representation is then passed to a classification module which predicts whether the corresponding sentence is a rationale (or not). A second model, using effectively the same architecture but parameterized independently, consumes the outputs (rationales) from the first to make predictions. This simple model is described at length in prior work~\citep{lehman2019inferring}. We further consider a `BERT-to-BERT' pipeline, where we replace each stage with a BERT module for prediction~\citep{devlin2018bert}.

In pipeline models, we train each stage independently. The rationale identification stage is trained using approximate sentence boundaries from our source annotations, with randomly sampled negative examples at each epoch. The classification stage uses the same positive rationales as the identification stage, a type of \emph{teacher forcing} \cite{williams1989learning} (details in Appendix~\ref{section:hyperparameters}).

\subsection{Soft selection}


We consider a model that passes tokens through BERT~\citep{devlin2018bert} to induce contextualized representations that are then passed to a bi-directional LSTM~\citep{hochreiter1997long}. The hidden representations from the LSTM are collapsed into a single vector using additive attention~\citep{bahdanau2014neural}. The LSTM layer allows us to bypass the 512 word limit imposed by BERT; when we exceed this, we effectively start encoding a `new' sequence (setting the positional index to 0) via BERT. The hope is that the LSTM learns to compensate for this. 
Evidence Inference and BoolQ comprise very long ($>$1000 token) inputs; we were unable to run BERT over these. 
We instead resorted to swapping GloVe 300d embeddings~\citep{pennington-etal-2014-glove} in place of BERT representations for tokens.

To soft score features we consider: Simple gradients, attention induced over contextualized representations, and LIME \cite{ribeiro2016should}.

\section{Evaluation}
\label{section:evaluation} 

\begin{table}
\scriptsize
    \centering
    \begin{tabular}{lccc}
    \toprule
         & Perf. & IOU F1 & Token F1 \\
    
    \midrule
    {\bf Evidence Inference} &  \\ 
        \citet{lei2016rationalizing} & 0.461 & 0.000 & 0.000 \\
    \citet{lei2016rationalizing} (u) & 0.461 & 0.000 & 0.000 \\
    \citet{lehman2019inferring} & 0.471 & 0.119 &  0.123 \\    
    Bert-To-Bert & 0.708 & 0.455 & 0.468 \\
        
    \midrule
    {\bf BoolQ} &  \\ 
    \citet{lei2016rationalizing} & 0.381  & 0.000  & 0.000 \\
    \citet{lei2016rationalizing} (u) & 0.380 & 0.000 & 0.000 \\
    \citet{lehman2019inferring} &  0.411  & 0.050 & 0.127 \\    
    Bert-To-Bert & 0.544 & 0.052 & 0.134 \\
    
     \midrule
    {\bf Movie Reviews} &  \\  
    \citet{lei2016rationalizing} & 0.914  &  0.124 &  0.285 \\
    \citet{lei2016rationalizing} (u) & 0.920 & 0.012 & 0.322 \\
    \citet{lehman2019inferring} & 0.750 & 0.063 & 0.139 \\
    Bert-To-Bert & 0.860 & 0.075 & 0.145 \\
    
    \midrule 
    {\bf FEVER} &  \\ 
    \citet{lei2016rationalizing} & 0.719 & 0.218 & 0.234  \\
    \citet{lei2016rationalizing} (u) & 0.718 & 0.000 & 0.000 \\
    \citet{lehman2019inferring}  & 0.691 & 0.540 &  0.523 \\    
    Bert-To-Bert & 0.877 & 0.835 & 0.812 \\
    
    \midrule
    {\bf MultiRC} &  \\ 
    \citet{lei2016rationalizing} &   0.655 &  0.271 &  0.456 \\
    \citet{lei2016rationalizing} (u) &  0.648 & 0.000$^\dagger$ & 0.000$^\dagger$ \\
    \citet{lehman2019inferring} & 0.614 & 0.136 & 0.140 \\
    Bert-To-Bert & 0.633 & 0.416 & 0.412 \\
    
    \midrule
    {\bf CoS-E} &  \\  
    \citet{lei2016rationalizing}   &  0.477 & 0.255  & 0.331  \\
    \citet{lei2016rationalizing}  (u)  &  0.476 & 0.000$^\dagger$  & 0.000$^\dagger$  \\
    Bert-To-Bert & 0.344 & 0.389 & 0.519\\
    
    \midrule
    {\bf e-SNLI} &  \\  
    \citet{lei2016rationalizing}   &  0.917 & 0.693 & 0.692 \\
    \citet{lei2016rationalizing}  (u)  &  0.903 & 0.261 & 0.379 \\
    Bert-To-Bert & 0.733 & 0.704 & 0.701 \\
    \bottomrule
    \end{tabular} 
    \caption{Performance of models that perform hard rationale selection. All models are supervised at the rationale level except for those marked with (u), which learn only from instance-level supervision; $^\dagger$ denotes cases in which rationale training degenerated due to the REINFORCE style training. Perf. is accuracy (CoS-E) or macro-averaged F1 (others). Bert-To-Bert for CoS-E and e-SNLI uses a token classification objective. Bert-To-Bert CoS-E uses the highest scoring answer.}
    \label{tab:hard-rationale-metrics}
\end{table}

\begin{table}[ht!]
\centering
\scriptsize
\begin{tabular}{lrrrr}
\toprule
         &  Perf. &  AUPRC &  Comp. $\uparrow$ &  Suff. $\downarrow$ \\
\midrule
\textbf{Evidence Inference} & & & & \\
         GloVe + LSTM - Attention &        0.429 &  0.506 &             -0.002 &       -0.023 \\
 GloVe + LSTM - Gradient &        0.429 &  0.016 &              0.046 &       -0.138 \\
  GloVe + LSTM - Lime &        0.429 &  0.014 &              0.006 &       -0.128 \\
  GloVe + LSTM - Random & 0.429 & 0.014 & -0.001 & -0.026 \\
  
  \midrule \textbf{BoolQ} & & & & \\ 
         GloVe + LSTM - Attention &        0.471 &  0.525 &              0.010 &        0.022 \\
 GloVe + LSTM - Gradient &        0.471 &  0.072 &              0.024 &        0.031 \\
 GloVe + LSTM - Lime &        0.471 & 0.073 &  0.028  &  -0.154 \\
 GloVe + LSTM - Random & 0.471 & 0.074 & 0.000 & 0.005 \\
 
 \midrule \textbf{Movies} & & & & \\
 BERT+LSTM - Attention &        0.970 &  0.417 &              0.129 &        0.097 \\
 BERT+LSTM - Gradient &        0.970 &  0.385 &              0.142 &        0.112 \\
 BERT+LSTM -           Lime &        0.970 &  0.280 &              0.187 &        0.093 \\
 BERT+LSTM - Random & 0.970 & 0.259  & 0.058 & 0.330 \\
 
 \midrule \textbf{FEVER} & & & & \\
 BERT+LSTM -        Attention &        0.870 &  0.235 &              0.037 &        0.122 \\
BERT+LSTM - Gradient &        0.870 &  0.232 &              0.059 &        0.136 \\
 BERT+LSTM -           Lime &        0.870 &  0.291 &              0.212 &        0.014 \\
 BERT+LSTM - Random & 0.870 & 0.244 & 0.034 & 0.122 \\
 
 \midrule \textbf{MultiRC} & & & & \\
 BERT+LSTM -        Attention &        0.655 &  0.244 &              0.036 &        0.052 \\
 BERT+LSTM - Gradient &        0.655 &  0.224 &              0.077 &        0.064 \\
 BERT+LSTM -           Lime &        0.655 &  0.208 &              0.213 &       -0.079 \\
 BERT+LSTM - Random & 0.655 & 0.186 & 0.029 & 0.081 \\
 
 \midrule \textbf{CoS-E} & & & & \\
 BERT+LSTM -        Attention &        0.487 &  0.606 &              0.080 &        0.217 \\
 BERT+LSTM - Gradient &        0.487 &  0.585 &              0.124 &        0.226 \\
 BERT+LSTM -           Lime &        0.487 &  0.544 &              0.223 &        0.143 \\
 BERT+LSTM - Random & 0.487 & 0.594  & 0.072 & 0.224 \\

\midrule \textbf{e-SNLI} & & & & \\
 BERT+LSTM -        Attention &        0.960 &  0.395 &              0.105 &        0.583 \\
 BERT+LSTM - Gradient &        0.960 &  0.416 &              0.180 &        0.472 \\
 BERT+LSTM -           Lime &        0.960 &  0.513 &              0.437 &        0.389 \\
 BERT+LSTM - Random & 0.960 & 0.357 & 0.081 & 0.487 \\
\bottomrule
\end{tabular}
\caption{Metrics for `soft' scoring models. Perf. is accuracy (CoS-E) or F1 (others). Comprehensiveness and sufficiency are in terms of AOPC (Eq. \ref{equation:AOPC}). `Random' assigns random scores to tokens to induce orderings; these are averages over 10 runs.}
\label{tab:soft-scores}
\vspace{-1em}
\end{table}

Here we present initial results for the baseline models discussed in Section \ref{section:baselines}, with respect to the metrics proposed in Section \ref{section:metrics}. 
We present results in two parts, reflecting the two classes of rationales discussed above: `Hard' approaches that perform discrete selection of snippets, and `soft' methods that assign continuous importance scores to tokens.

In Table \ref{tab:hard-rationale-metrics} we evaluate models that perform discrete selection of rationales. We view these as inherently faithful, because by construction we know which snippets the decoder used to make a prediction.\footnote{This assumes independent encoders and decoders.} Therefore, for these methods we report only metrics that measure agreement with human annotations. 

Due to computational constraints, we were unable to run our BERT-based implementation of~\citet{lei2016rationalizing} over larger corpora. Conversely, the simple pipeline of \citet{lehman2019inferring} assumes a setting in which rationale are sentences, and so is not appropriate for datasets in which rationales tend to comprise only very short spans. Again, in our view this highlights the need for models that can rationalize at varying levels of granularity, depending on what is appropriate. 

We observe that for the ``rationalizing'' model of~\citet{lei2016rationalizing}, exploiting rationale-level supervision often (though not always) improves agreement with human-provided rationales, as in prior work \citep{zhang2016rationale,strout2019human}. Interestingly, this does not seem strongly correlated with predictive performance. 

~\citet{lei2016rationalizing} outperforms the simple pipeline model when using a BERT encoder. Further, ~\citet{lei2016rationalizing} outperforms the `BERT-to-BERT' pipeline on the comparable datasets for the final prediction tasks. This may be an artifact of the amount of text each model can select: `BERT-to-BERT' is limited to sentences, while ~\citet{lei2016rationalizing} can select any subset of the text.
Designing extraction models that learn to adaptively select contiguous rationales of appropriate length for a given task seems a potentially promising direction.



In Table \ref{tab:soft-scores} we report metrics for models that assign continuous importance scores to individual tokens. For these models we again measure downstream (task) performance (macro F1 or accuracy). Here the models are actually the same, and so downstream performance is equivalent. To assess the quality of token scores with respect to human annotations, we report the Area Under the Precision Recall Curve (AUPRC). 

These scoring functions assign only soft scores to inputs (and may still use all inputs to come to a particular prediction), so we report the metrics intended to measure faithfulness defined above: comprehensiveness and sufficiency, averaged over `bins' of tokens ordered by importance scores. To provide a point of reference for these metrics --- which depend on the underlying model --- we report results when rationales are randomly selected (averaged over 10 runs).

Both simple gradient and LIME-based scoring yield more comprehensive rationales than attention weights, consistent with prior work \cite{jain2019attention,serrano2019attention}. Attention fares better in terms of AUPRC --- suggesting better agreement with human rationales --- which is also in line with prior findings that it may provide plausible, but not faithful, explanation \cite{zhong2019fine}.
Interestingly, LIME does particularly well across these tasks in terms of faithfulness. 

From the `Random' results that we conclude models with overall poor performance on their final tasks tend to have an overall poor ordering, with marginal differences in comprehensiveness and sufficiency between them. For models that with high sufficiency scores: Movies, FEVER, CoS-E, and e-SNLI, we find that random removal is particularly damaging to performance, indicating poor absolute ranking; whereas those with high comprehensiveness are sensitive to rationale length.






\section{Conclusions and Future Directions}

We have introduced a new publicly available resource: the Evaluating Rationales And Simple English Reasoning (ERASER) benchmark. This comprises seven datasets, all of which include both instance level labels and corresponding supporting snippets (`rationales') marked by human annotators. We have augmented many of these datasets with additional annotations, and converted them into a standard format comprising inputs, rationales, and outputs.
ERASER is intended to facilitate progress on explainable models for NLP. 

We proposed several metrics intended to measure the quality of rationales extracted by models, both in terms of agreement with human annotations, and in terms of `faithfulness'. 
We believe these metrics provide reasonable means of comparison of specific aspects of interpretability, but we view the problem of measuring faithfulness, in particular, a topic ripe for additional research (which ERASER can facilitate). 

Our hope is that ERASER enables future work on designing more interpretable NLP models, and comparing their relative strengths across a variety of tasks, datasets, and desired criteria. It also serves as an ideal starting point for several future directions such as better evaluation metrics for interpretability, causal analysis of NLP models and datasets of rationales in other languages.

%

\section{Acknowledgements}
We thank the anonymous ACL reviewers.

This work was supported in part by the NSF (CAREER award 1750978), and by the Army Research Office (W911NF1810328). 


\bibliography{acl2020}
\bibliographystyle{acl_natbib}


\appendix
\begin{center}
    {\large {\bf Appendix} }

\end{center}

\section{Dataset Preprocessing}\label{section:preprocessing}

\begin{table*}
\centering
\footnotesize
\begin{tabular}{lrrrrr}
\toprule
            Dataset &  Documents &  Instances & Rationale \% &  Evidence Statements & Evidence Lengths \\
\midrule
 \textbf{MultiRC} \\ 
 Train &  400 &  24029 &  17.4 &  56298 &  21.5 \\
 Val &  56 &  3214 &  18.5 &  7498 &  22.8 \\
 Test &  83 &  4848 & -  & - &  - \\\hline
 \textbf{Evidence Inference} \\ 
 Train &  1924 &  7958 &  1.34 &  10371 &  39.3 \\
 Val &  247 &  972 &  1.38 &  1294 &  40.3 \\
 Test &  240 &  959 &  - & - & -  \\\hline
 \textbf{Exhaustive Evidence Inference} \\ 
 Val & 81 &  101 &  4.47 &  504.0 &  35.2 \\
 Test &  106 &  152 &  - & - & -  \\\hline
 \textbf{Movie Reviews} \\
 Train &  1599 &  1600 &  9.35 &  13878 &  7.7 \\
 Val & 150 &  150 &  7.45 &  1143.0 &  6.6 \\
 Test &  200 &  200 &  - & - &  - \\\hline
 \textbf{Exhaustive Movie Reviews} \\
 Val & 50 &  50 &  19.10 &  592.0 &  12.8 \\
 \textbf{FEVER} \\
 Train &  2915 &  97957 &  20.0 &  146856 &  31.3 \\
 Val &  570 &  6122 &  21.6 &  8672 &  28.2 \\
 Test &  614 &  6111 & -  & - &  - \\\hline
 \textbf{BoolQ} \\
  Train &  4518 &  6363 &  6.64 &  6363.0 &  110.2 \\
 Val &  1092 &  1491 &  7.13 &  1491.0 &  106.5 \\
 Test &  2294 &  2817 & - & - &  - \\\hline
 \textbf{e-SNLI} \\
 Train &  911938 &  549309 &  27.3 &  1199035.0 &  1.8 \\
 Val &  16328 &  9823 &  25.6 &  23639.0 &  1.6 \\
 Test &  16299 &  9807 & - & - &  - \\\hline
 \textbf{CoS-E} \\
 Train&  8733 &  8733 &  26.6 &  8733 &  7.4 \\
 Val &  1092 &  1092 &  27.1 &  1092 &  7.6 \\
 Test &  1092 &  1092 &  - & - &  - \\
\bottomrule
\end{tabular}
\caption{Detailed breakdowns for each dataset - the number of documents, instances, evidence statements, and lengths. Additionally we include the percentage of each relevant document that is considered a rationale. For test sets, counts are for all instances including documents with non comprehensive rationales.}
\end{table*}\label{table:detailed_dataset}

\begin{table*}
\centering
\scriptsize
\begin{tabular}{lccccc}
\toprule
Dataset &  Labels &  Instances &  Documents & Sentences &  Tokens \\
\midrule
 Evidence Inference &  3 &  9889 &  2411 &  156.0 &  4760.6 \\
 BoolQ &  2 &  10661 &  7026 &  175.3 &  3582.5 \\
 Movie Reviews &  2 &  2000 &  1999 &  36.8 &  774.1 \\
 FEVER &  2 &  110190 &  4099 &  12.1 &  326.5 \\
 MultiRC &  2 &  32091 &  539 &  14.9 &  302.5 \\
 CoS-E &  5 &  10917 &  10917 &  1.0 &  27.6 \\
 e-SNLI &  3 &  568939 &  944565 &  1.7 &  16.0 \\
\bottomrule
\end{tabular}
\caption{General dataset statistics: number of labels, instances, unique documents, and average numbers of sentences and tokens in documents, across the publicly released train/validation/test splits in ERASER. For CoS-E and e-SNLI, the sentence counts are not meaningful as the partitioning of question/sentence/answer formatting is an arbitrary choice in this framework.}
\end{table*}\label{table:generic_dataset_information}

We describe what, if any, additional processing we perform on a per-dataset basis. All datasets were converted to a unified format.

\vspace{.5em} \noindent {\bf MultiRC} \citep{KCRUR18-multirc} We perform minimal processing. We use the validation set as the testing set for public release.

\vspace{.5em} \noindent {\bf Evidence Inference} \citep{lehman2019inferring} We perform minimal processing. As not all of the provided evidence spans come with offsets, we delete any prompts that had no grounded evidence spans.

\vspace{.5em} \noindent {\bf Movie reviews} \citep{zaidan2008modeling} We perform minimal processing. We use the ninth fold as the validation set, and collect annotations on the tenth fold for comprehensive evaluation.

\vspace{.5em} \noindent {\bf FEVER} \citep{thorne2018fever} We perform substantial processing for FEVER - we delete the "Not Enough Info" claim class, delete any claims with support in more than one document, and repartition the validation set into a validation and a test set for this benchmark (using the test set would compromise the information retrieval portion of the original FEVER task). We ensure that there is no document overlap between train, validation, and test sets (we use \citet{pearce2005improved} to ensure this, as conceptually a claim may be supported by facts in more than one document). We ensure that the validation set contains the documents used to create the FEVER symmetric dataset \citep{shuster-symmetric} (unfortunately, the documents used to create the validation and test sets overlap so we cannot provide this partitioning). Additionally, we clean up some encoding errors in the dataset via \citet{speer-2019-ftfy}.

\vspace{.5em} \noindent {\bf BoolQ} \citep{clark2019boolq} The BoolQ dataset required substantial processing. The original dataset did not retain source Wikipedia articles or collection dates. In order to identify the source paragraphs, we download the 12/20/18 Wikipedia archive, and use FuzzyWuzzy \url{https://github.com/seatgeek/fuzzywuzzy} to identify the source paragraph span that best matches the original release. If the Levenshtein distance ratio does not reach a score of at least 90, the corresponding instance is removed. For public release, we use the official validation set for testing, and repartition train into a training and validation set.

\vspace{.5em} \noindent {\bf e-SNLI}~\citep{camburu2018snli} We perform minimal processing. We separate the premise and hypothesis statements into separate documents.

\vspace{.5em} \noindent {\bf Commonsense Explanations (CoS-E)} \citep{rajani2019explain} We perform minimal processing, primarily deletion of any questions without a rationale or questions with rationales that were not possible to automatically map back to the underlying text. As recommended by the authors of \citet{talmor-etal-2019-commonsenseqa} we repartition the train and validation sets into a train, validation, and test set for this benchmark. We encode the entire question and answers as a prompt and convert the problem into a five-class prediction. We also convert the ``Sanity" datasets for user convenience.

All datasets in ERASER were tokenized using spaCy\footnote{\url{https://spacy.io/}} library (with SciSpacy \citep{Neumann2019ScispaCyFA} for Evidence Inference). In addition, we also split all datasets except e-SNLI and CoS-E into sentences using the same library.

\section{Annotation details}
\label{section:ann-details}
We collected \emph{comprehensive} rationales for a subset of some test sets to accurately evaluate model recall of rationales.

\begin{enumerate}
    
    \item \textbf{Movies}. We used the Upwork Platform\footnote{\url{http://www.upwork.com}} to hire two fluent english speakers to annotate each of the 200 documents in our test set. Workers were paid at rate of USD 8.5 per hour and on average, it took them 5 min to annotate a document. Each annotator was asked to annotate a set of 6 documents and compared against in-house annotations (by authors).

    \item \textbf{Evidence Inference}. We again used Upwork to hire 4 medical professionals fluent in english and having passed a pilot of 3 documents. 125 documents were annotated (only once by one of the annotators, which we felt was appropriate given their high-level of expertise) with an average cost of USD 13 per document. Average time spent of single document was 31 min. 
    
    \item \textbf{BoolQ}. We used Amazon Mechanical Turk (MTurk) to collect reference comprehensive rationales from randomly selected 199 documents from our test set (ranging in 800 to 1500 tokens in length). Only workers from AU, NZ, CA, US, GB with more than 10K approved HITs and an approval rate of greater than 98\% were eligible. For every document, 3 annotations were collected and workers were paid USD 1.50 per HIT. The average work time (obtained through MTurk interface) was 21 min. We did not anticipate the task taking so long (on average); the effective low pay rate was unintended.
    
\end{enumerate}

\section{Hyperparameter and training details}\label{section:hyperparameters}

\subsection{\citep{lei2016rationalizing} models}
For these models, we set the sparsity rate at 0.01 and we set the contiguity loss weight to 2 times sparsity rate (following the original paper). We used bert-base-uncased ~\citep{Wolf2019HuggingFacesTS} as token embedder (for all datasets except BoolQ, Evidence Inference and FEVER) and Bidirectional LSTM with 128 dimensional hidden state in each direction. A dropout~\citep{JMLR:v15:srivastava14a} rate of 0.2 was used before feeding the hidden representations to attention layer in decoder and linear layer in encoder. One layer MLP with 128 dimensional hidden state and ReLU activation was used to compute the decoder output distribution. 

For three datasets mentioned above, we use GloVe embeddings (\url{http://nlp.stanford.edu/data/glove.840B.300d.zip}).

A learning rate of 2e-5 with Adam~\citep{Kingma2014AdamAM} optimizer was used for all models and we only fine-tuned top two layers of BERT encoder. Th models were trained for 20 epochs and early stopping with patience of 5 epochs was used. The best model was selected on validation set using the final task performance metric.

The input for the above model was encoded in form of \texttt{[CLS] document [SEP] query [SEP]}.

This model was implemented using the {\tt AllenNLP} library~\citep{Gardner2017AllenNLP}.

\subsection{BERT-LSTM/GloVe-LSTM}

This model is essentially the same as the decoder in previous section. The BERT-LSTM uses the same hyperparameters, and GloVe-LSTM is trained with a learning rate of 1e-2.

\subsection{\citet{lehman2019inferring} models}

With the exception of the Evidence Inference dataset, these models were trained using the GLoVe ~\citep{pennington-etal-2014-glove} 200 dimension word vectors, and Evidence Inference using the ~\citep{Pyysalo2013DistributionalSR} PubMed word vectors. We use Adam ~\citep{Kingma2014AdamAM} with a learning rate of 1e-3, Dropout~\citep{JMLR:v15:srivastava14a} of 0.05 at each layer (embedding, GRU, attention layer) of the model, for 50 epochs with a patience of 10. We monitor validation loss, and keep the best model on the validation set.

\subsection{BERT-to-BERT model}

We primarily used the `bert-base-uncased` model for both components of the identification and classification pipeline, with the sole exception being Evidence Inference with SciBERT ~\citep{Beltagy2019SciBERT}. We trained with the standard BERT parameters of a learning rate of 1e-5, Adam ~\citep{Kingma2014AdamAM}, for 10 epochs.  We monitor validation loss, and keep the best model on the validation set.



\end{document}